\pdfoutput=1

\documentclass[11pt]{article}

\PassOptionsToPackage{numbers}{natbib}
\usepackage[preprint]{acl}

\usepackage{hyperref}
\usepackage{times}
\usepackage{latexsym}
\usepackage{multirow} 
\usepackage[T1]{fontenc}

\usepackage[utf8]{inputenc}
\usepackage{algorithm}
\usepackage{algpseudocode}
\usepackage{subcaption}
\usepackage{microtype}

\usepackage{inconsolata}

\usepackage{graphicx}
\usepackage{multicol}

\usepackage{amsmath}
\usepackage{amssymb}
\usepackage{mathtools}
\usepackage{amsthm}
\usepackage{cleveref}
\usepackage{url}            %
\usepackage{booktabs}       %
\usepackage{amsfonts}       %
\usepackage{nicefrac}       %
\usepackage{color}
\usepackage{xcolor}
\usepackage{natbib}
\usepackage{dsfont}

\DeclareMathOperator*{\argmax}{arg\,max}

\title{Jump Starting Bandits with LLM-Generated Prior Knowledge}

\author{Parand A. Alamdari\thanks{Work performed during an internship at Borealis AI.} \\
  University of Toronto \\ \& Vector Institute \\
  \texttt{parand@cs.toronto.edu} \\\And
  Yanshuai Cao \\
  Borealis AI \\
  \hspace{5.3cm}\texttt{\{yanshuai.cao, kevin.h.wilson\}@borealisai.com}\\\And
   Kevin H. Wilson \\
  Borealis AI\\
  }

\begin{document}
\renewcommand{\le}{\leqslant}
\renewcommand{\leq}{\leqslant}
\renewcommand{\ge}{\geqslant}
\renewcommand{\geq}{\geqslant}

\renewcommand{\bar}{\overline}
\renewcommand{\hat}{\widehat}

\newcommand{\IR}{\mathbb{R}}
\newcommand{\NN}{\mathbb{N}}

\newcommand{\llm}{\mathcal{M}}
\newcommand{\llmout}{\mathcal{A}}

\newcommand{\expectedvalue}{\mathbb{E}}

\newcommand{\avg}{\operatorname{avg}}

\newcommand{\reg}{\operatorname{reg}}

\maketitle

\begin{abstract}
We present substantial evidence demonstrating the benefits of integrating Large Language Models (LLMs) with a Contextual Multi-Armed Bandit framework. Contextual bandits have been widely used in recommendation systems to generate personalized suggestions based on user-specific contexts. We show that LLMs, pre-trained on extensive corpora rich in human knowledge and preferences, can simulate human behaviours well enough to jump-start contextual multi-armed bandits to reduce online learning regret. We propose an initialization algorithm for contextual bandits by prompting LLMs to produce a pre-training dataset of approximate human preferences for the bandit. This significantly reduces online learning regret and data-gathering costs for training such models. Our approach is validated empirically through two sets of experiments with different bandit setups: one which utilizes LLMs to serve as an oracle and a real-world experiment utilizing data from a conjoint survey experiment.
\end{abstract}

\section{Introduction}
When users encounter content that caters to their needs and preferences, they are more inclined to engage, e.g., watch a movie \cite{amat2018artwork} or click on a news article \cite{li2010contextual}. But how should a practitioner learn to personalize a campaign's content? Would a user be more likely to respond if the content was presented in a formal or informal style \cite{linos2024formality}? Or if it included a celebrity endorsement \cite{kreps2020factors}? Studies in behavioural economics \cite{thaler2021nudge, epstein2022can} proved that the answers can be heavily context-dependent \cite{dai2021behavioural, rabb2022evidence}. 

Contextual Multi-Armed Bandits (CBs) \cite{lu2010contextual, chu2011contextual} were invented to address this problem in the online learning setting---where an agent with no prior knowledge of user preferences is presented with users in a sequence, chooses a piece of content to show the users based on the users' and the content's features, and updates itself based on feedback essentially instantaneously. The critical question for CBs is how to balance {\em exploration} (gathering information about users' preferences) and {\em exploitation} (utilizing the information it has gathered so far). While these agents are known to exhibit good asymptotic performance, their initial choices are essentially random. To improve performance for the first users in a campaign, researchers have focused on {\em warm starting} a CB \cite{zhang2019warm} using detailed records of users' past behaviours and preferences in similar campaigns. However, collecting such datasets poses significant challenges due to resource demands, data diversity requirements, and the need to comply with privacy regulations. 

In this paper, we show that large language models (LLMs) can solve this conundrum by {\em jump starting} contextual bandits using approximate human preference information already captured in them. This allows practitioners to automatically design and test context-dependent messaging with significantly lower upfront costs. 

Prior work has shown conflicting evidence about whether LLMs faithfully simulate human behaviours. \citet{argyle2023out} and \citet{simmons2022moral} show LLMs can simulate human behaviour and preferences in certain circumstances, while \citet{santurkar2023whose} find that LLM-generated response distributions may not match the true population distribution. 
Our {\em key insight} is that even if LLM-generated reward distributions do not perfectly match human preferences, they still provide a {\em better than random} baseline than that which cold-started CBs face.

{\bf Our contributions.}
First, we introduce a novel approach, Contextual Bandits with LLM Initialization (CBLI), that leverages the power of LLMs to generate synthetic reward distributions for pretraining CBs for personalization in \Cref{alg:CBLI}.

Second, we empirically demonstrate the effectiveness of our proposed method to pre-train a CB in two settings where natural language plays a critical role in the content shown to users. The first setting involves a standard contextual bandit whose arms represent the styles of marketing communication soliciting charity donations from simulated users. The second experiment uses a {\em sleeping bandit} setup with real-world human preferences. CBLI achieves 14--17\% and 19--20\% reduction in early regret, respectively. Furthermore, we show that even when certain privacy-sensitive attributes are withheld, CBLI still achieves an 14.8\% reduction in early regret in the second setting.

Finally, testing contextual bandit algorithms in real-world settings usually requires collecting a large amount of randomized log data, which can be costly and difficult while maintaining data diversity and privacy requirements. As an additional contribution, we introduce, to the ML and NLP community, the usage of conjoint survey experimental data \cite{Bansak2021} as a way to benchmark bandit algorithms. Conjoint experiment data collected in social science literature provides a substantial source of real-world data where users were presented with several potential choices and asked to record their preferences. Coupled with the demographic information that is typically available from such experiments from which contexts can be constructed, they are an ideal candidate for testing contextual bandit algorithms.

\section{Related Work}

With the advent of web scale content, recommender systems that personalized content to specific users became an important part of navigating the internet. Recently, deep neural networks have become more important to these systems \cite{cheng2016wide, covington2016deep}, and the need to train these systems online and at scale has been a prominent concern \cite{liu2022monolith}.

One such online training technique, multi-armed bandits, has proven especially popular, with a plethora of algorithms specialized to specific settings (see \cite{lattimore2020bandit, slivkins2019introduction} for an introduction). Most relevant to this work are contextual bandits \cite{wang2005bandit, pandey2007bandits, NIPS2007_4b04a686}, which introduce side information to determine the optimal arm. In particular, \citet{zhang2019warm} consider warm starting CBs utilizing fully supervised, labelled data collected from human experts.

While to the best of our knowledge, conjoint survey experiments have not be used in the evaluation of contextual bandits, conjoint analysis---the underlying statistical technique---has been used for feature selection in training CBs \cite{li2010contextual}. Outside of CBs, researchers have used ML techniques to improve the efficiency of conjoint analysis \cite{chapelle2004machine, hainmueller2014causal, ham2022using}.

Recently, researchers have attempted to integrate LLMs into social science studies, specifically utilizing LLMs to simulate human behaviours in social psychology experiments \cite{dillion2023can, aher2023using} and predict their opinions \cite{argyle2023out, simmons2022moral}. Additionally, a growing body of research has focused on evaluating LLM performance and biases through survey questions. Prior studies have examined how LLMs respond to multiple-choice questions from public opinion polls, revealing insights into their alignment with human responses and potential biases \cite{santurkar2023whose, durmus2023towards, dominguez2023questioning}. \citet{santurkar2023whose} focus on diverse demographic groups across the U.S., while \citet{durmus2023towards} focus on non-American demographics, both finding that LLMs' responses to surveys show discrepancies with the collected answers from representative samples. Furthermore, \citet{dominguez2023questioning} show that models' responses are sensitive to ordering and labelling biases, and they do not contain the statistical signals typically found in human populations. 

Our work in this setting suggests that \emph{despite} potential biases and disparities in LLM-generated responses, they provide a good starting point for training contextual multi-armed bandits.

\section{Preliminaries}
\label{sec:prelim}

\paragraph{Multi-Armed Bandits.}
In a multi-armed bandit problem, an agent, at each step $t \in [T]$, is presented with a set of $K$ possible \emph{arms} and must choose an arm $k_t$. After the agent chooses, they receive a \emph{reward} $r_{t, k_t}$ drawn from a {\em reward distribution} $R(k_t)$, which is initially unknown. We focus on the {\em stochastic} setting, where each $r_{t, k}$ is sampled independently from $R(k)$. The goal of the agent is to maximize the total reward $\sum_{t = 1}^T r_{t, k_t}$. To achieve this, the agent needs to balance the trade-off between exploration (i.e., trying different arms to gather more information about their reward distribution) and exploitation (i.e., choosing among the arms that appear to offer the highest reward based on the collected information so far).

\paragraph{Contextual Multi-Armed Bandits.} 
In contextual multi-armed bandits (CB), where at each time step $t \in [T]$, the agent receives side information $\phi_t \in \mathcal{C}$ which is called the {\em context}. The reward distribution of pulling arm $k \in [K]$ at time $t$ may depend on the context $\phi_t$, that is, $r_{t, k} \sim R(k, \phi_t)$

\paragraph{Policy.} A policy $\pi$ is a map from histories $H_t = (\phi_1, k_1, r_{1, k_1}, \ldots, \phi_{t-1}, k_{t-1}, r_{t-1, k_{t-1}}, \phi_t)$ of contexts, arms, and realized rewards to the next arm $\pi(H_t) = k$ the agent will choose.

\paragraph{Regret.} The {\em (cumulative) regret} of an agent after $T$ steps choosing arms based on a policy $\pi$ relative to a policy $\pi^\ast$ is given by

$$
  \reg_{\pi, \pi^\ast} = 
    \expectedvalue 
      \left[ \sum_{t=1}^T r_{t, k^*_t} \right] 
    - \expectedvalue
      \left[ \sum_{t=1}^T r_{t, k_t} \right]
$$  

\noindent
where $k_t$ is chosen according to $\pi$ and $k^\ast_t$ is chosen according to $\pi^\ast$ and the expectations are taken with respect to any randomness in the rewards, contexts, and policy.

\paragraph{Sleeping bandit.} 
In some of our applications, we will consider {\em sleeping bandits}, where at time $t$, the agent may choose between a subset of arms $\mathcal{A}_t \subseteq [K]$ \cite{kleinberg2010regret}. In this case, the bandit algorithm needs to learn and determine which arms yield the best reward {\em when they are available}.

\section{Method}
\label{sec:method}

Recent advances in large language models have shown remarkable progress in understanding and interpreting human expression \cite{zhao2023survey}. LLMs trained on extensive corpora rich in human knowledge preserve the capacity to perform well on a diverse array of tasks, even those dependent on human behaviour and preferences \cite{brown2020language}. In the context of personalization, LLMs offer two key advantages. First, they can be utilized to automatically design user-specific content in large volumes. 
Second, LLMs can be used to simulate human interactions and predict their preferences, effectively serving as a proxy for collecting a dataset of users' interactions. We focus on the latter in our work, and provide a framework to utilize the power of LLMs to generate extensive artificial user interactions to later train our models.  

\subsection{Problem Formulation}
\label{sec:formulation}

We showcase our proposed approach for two different bandit setups: the standard contextual multi-armed bandit (CB) and the sleeping bandit \cite{kleinberg2010regret} introduced in \Cref{sec:prelim}.

We consider a set of $n$ users, where the vector of features of each user $i \in [n]$ is denoted by $X_i$, which is sampled from a population $\mathcal{X}$. At each time step $t$, user $u_t$ is sampled from the set of $n$ users. The user's {\em context} can be calculated using a mapping function $\phi: \mathcal{X} \rightarrow \mathcal{C}$ where $\mathcal{C}$ is the context space. There are $K$ different arms, each representing a potential recommendation for the user. For each user $u_t$ and arm $k$, there is a scalar reward $R(k, u_t)$ following some unknown distribution. For brevity, we write $R_{t}$ henceforth. We note that there is an optimal policy $\pi_o$ which chooses the optimal arm $k^\ast_t = \argmax_{k \in [K]} \expectedvalue[R_t(k)]$ at each time step. We seek to optimize the following (expected, cumulative) regret:

\begin{align*}
    \pi^* = \arg\min_{\pi} \reg_{\pi, \pi_o}
\end{align*}

Following the standard treatment in stochastic contextual bandits \cite{lattimore2020bandit}, we use a feature map $\psi\!\!:\!\mathcal{C}\! \times\! [K]\! \rightarrow \!\mathbb{R}^{d}$ that jointly encodes context and arms (either all arms or just the available ones in sleeping bandits). Then we model reward using $R_t = \langle \psi_t, \theta \rangle$, where $\theta$ is the parameter to be learned over the course of $T$ steps. In our experiments, we adapt the LinUCB algorithm \cite{chu2011contextual} in order to train these bandits. From a cold start, this algorithm is guaranteed to yield regret of $\widetilde{O}(\sqrt{Td})$ where $d$ is the dimension of the vector $\theta$.

\begin{figure*}[t]
\centering
\begin{subfigure}[t]{0.49\textwidth}
\includegraphics[width=\linewidth]{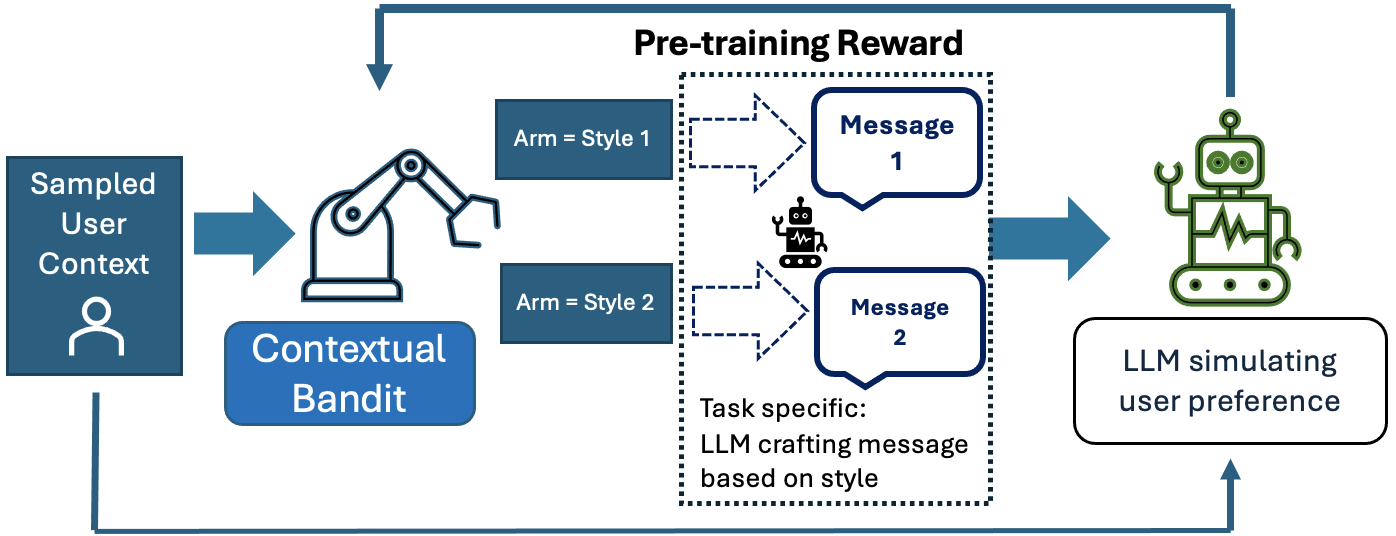}
\end{subfigure}
\begin{subfigure}[t]{0.49\textwidth}
\includegraphics[width=\linewidth]{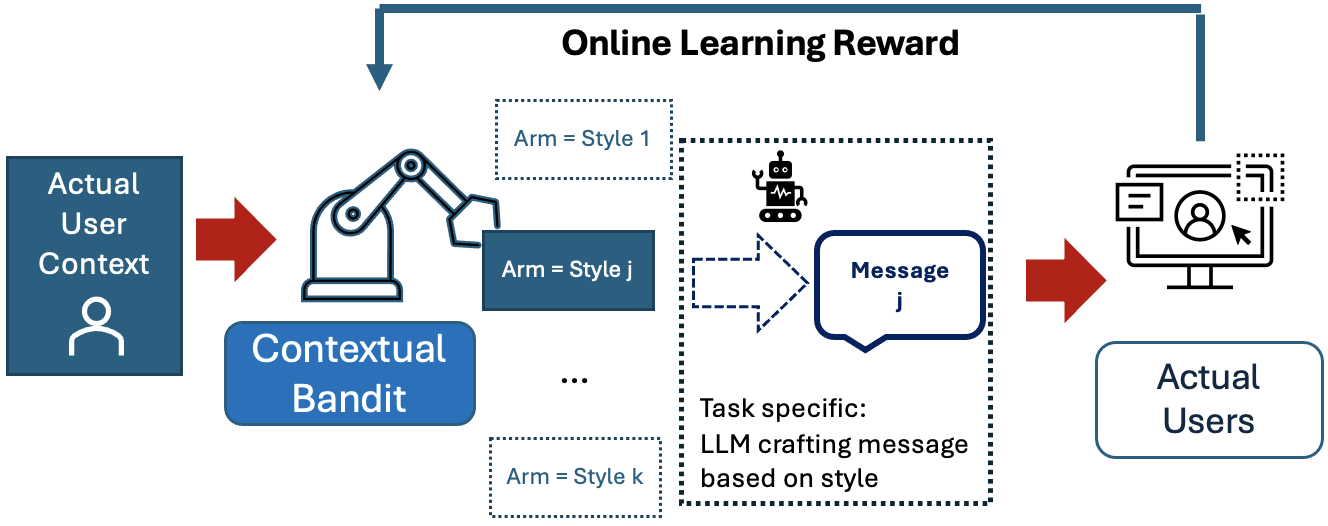}
\end{subfigure}
\caption{Using LLM to jump-start bandit learning: (left) pre-training using our proposed CBLI; (right) online learning of jump-started bandit.}
    \label{figs:banditprompt}
\end{figure*}

We introduce our framework ``Contextual Bandits with LLM Initialization (CBLI)''. We propose to use LLMs to generate a large and diverse dataset of users' interactions and their preferences. We use this dataset to pretrain a contextual bandit which we then use in a real setting. The trained model serves as a starting point, and later is tuned with the data of real users' interactions as they are collected. The overall framework is depicted in \Cref{figs:banditprompt}.

\subsection{Contextual Bandit with LLM Initialization (CBLI)}

\begin{algorithm}[t!]
\caption{Generate Synthetic Preferences}
\label{alg:CBLI}
\begin{algorithmic}[1]
\State {\bfseries Input:} Specification of features of users, $K$ Arms, number of users $n$, an LLM $\llm$, and {is\_sparse\_mode}
\State $ U \leftarrow \{\}$, $R \leftarrow \{\}$
\For{$i \in [n]$}
\State  Sample user $i$ i.i.d.~from feature space $\mathcal{X}$
\State Embed user $i$ to the context space as $\phi_i$ 
\State Describe user $i$ with text as $\mathcal{T}_i$
\State $U \leftarrow U \cup \{ \phi_i\}$ 
\If {is\_sparse\_mode}
\State $R_i \leftarrow [0]^{K \times K}$ \Comment{sparse matrix}
\State $A = \text{Uniform}([K] \times [K])$
\Else
\State $R_i \leftarrow [0]^K$ \Comment{vector}
\State $A = [K] \times [K]$
\EndIf
\State Prompt $\llm$ to adopt $i$'s persona given $\mathcal{T}_i$
 
\For{$(k_1,k_2) \in A $}
\State Prompt $\llm$ to indicate user $i$'s preference between $k_1$ and $k_2$ as $P$
\If {is\_sparse\_mode}
\State $R_i[k_1, k_2] +\!\!= \mathds{1}[k_1 == P]$ 
\Else
\State $R_i[P] +\!\!= 1$
\EndIf
\EndFor
\State Normalize $R_i$
\State $R \leftarrow R \cup \{ R_i\}$
\EndFor
\State {\bfseries Output:} Users' Context: $U$, Rewards: $R$
\end{algorithmic}
\end{algorithm}

We begin by setting up a contextual bandit framework where the context is the information about each user as described in \Cref{sec:formulation}. First, we generate $n$ synthetic users by sampling i.i.d.~from $\mathcal{X}$. For each sampled user $i$, whose features we denote by $X_i$, we compute a textual embedding $\mathcal{T}_i = \mathcal{T}(X_i) \in \mathcal{L}$ in some language space $\mathcal{L}$. The exact function $\mathcal{T}$ is domain-dependent and can represent arbitrary side information about a user. For instance, a video streaming service might take $X_i$ to be the sequence of movies the user $i$ has watched and transform it into the string ``This user has watched the following movies: [movie$_1$], [movie$_2$], \ldots'' These textual embeddings are then transformed into a context $\phi_i = \phi(\mathcal{T}_i)$ utilizing another LLM.
 
The $K$ arms of the bandit represent the different options we might offer to the users or, as discussed in \Cref{sec:experiments-synthetic}, potential prompts to an LLM which will generate content to send to users.

To estimate the reward distribution for pulling each arm, given a user's context, we propose the following approach. For each generated user $i$, we use an LLM $\llm$ to simulate the preferences of user $i$ based on their textual representation $\mathcal{T}_i$. Specifically, for each user $i$, we prompt $\llm$ to adopt the persona of user $i$. To determine the reward distribution for different actions, we consider each pair of arms $(k_1, k_2)$ within the set of $K$ arms. We then prompt $\llm$ to indicate which arm user $i$ would prefer. See \Cref{sec:appendix-experimental-details} for the prompts utilized. We repeat the process multiple times to get more instances of answers for each pair of arms and user. The algorithm is described in \Cref{alg:CBLI}. 

In the case of a sleeping bandit or when the number of arms is large, we can use the sparse mode in \Cref{alg:CBLI}, where only a random subset of pairs are sampled, and pairwise preferences are stored. Alternatively, \Cref{alg:CBLI} iterates over all pairs of arms and records an absolute reward per each arm based on the number of winnings in pairwise comparisons.   

We note three features of CBLI. First, we prompt $\llm$ to rank {\em pairs} of arms as opposed to scoring arms individually (see \Cref{app:scoring} for more details). Second, we run over all pairs $(k_1, k_2) \in [K] \times [K]$ (Lines 10 and 13) and not, say, all pairs $(k_1, k_2)$ with $k_1 < k_2$, which should be sufficient to determine an ordering. However, it has been observed that LLMs are sensitive to the order options are presented \cite{santurkar2023whose}, so we average over both orders to mitigate this potential bias. Third, we note that $R_i[k]$ should approximate a {\em rank ordering} of per arm rewards and may not estimate the exact reward (see \Cref{sec:appendix_rank_order}), which is sufficient for our problem of best arm identification, though see \Cref{sec:limitations} for further discussion.

We leverage the dataset of generated users and their corresponding rewards to pre-train a contextual multi-armed bandit using established algorithms like those described in \cite{lu2010contextual, chu2011contextual}. At each step $t \in [T]$, we sample a user $u_t$ and their associated context $\phi_{t}$ from the set of $n$ generated users. In the usual setting, the bandit then chooses an arm $a_t$ and receives reward $R_t[k_t]$. In the sleeping bandit setting, the bandit will be presented with two arms $(k_1, k_2)$ and will choose one to set as $k_1$ and the other as $k_2$. It will then receive reward $R_t[k_1, k_2]$. We note that we always have access to (the CBLI-generated) $R_{t}$ unlike training from actual log data, where the user $u_t$ may not have been assigned to the desired arm.

In \Cref{sec:experiments} we show that the pre-trained bandit model using LLM generated users and rewards substantially reduces regret in decision-making. More importantly, utilizing LLM-generated user data and their simulated preferences offers significant benefits. It not only lowers the costs associated with data collection but also mitigates privacy concerns, as it does not involve real user data. CBLI provides a cost-effective and privacy-preserving solution for training models in personalized systems.

\section{Experiments}
\label{sec:experiments}
We validate the effectiveness of our proposed approach through two sets of experiments.\footnote{Code for both experiments is available at \href{https://github.com/BorealisAI/jump-starting-bandits}{https://github.com/BorealisAI/jump-starting-bandits}.} In both sets of experiments, we adapt the LinUCB algorithm \cite{chu2011contextual} to train our bandit. The first set of experiments is based on synthetic data, and the second set uses a real-world dataset derived from a choice-based conjoint analysis \cite{DVN/6BSJYP_2020}. Experimental details are in \Cref{sec:appendix-experimental-details}.  

\paragraph{Language Models.} To evaluate the effectiveness of our approach, we configure our model $\llm$ in \Cref{alg:CBLI} to several state-of-the-art Large Language Models: OpenAI GPT-4o \cite{achiam2023gpt}, GPT-3.5, Claude-3 Haiku \cite{claude3}, and Mistral-Small \cite{jiang2023mistral}.

\subsection{First Experiment: Personalized Email Campaign for Charity Donations}
\label{sec:experiments-synthetic}

To verify the effectiveness of using LLMs to pre-train contextual multi-armed bandits, we design an experiment focused on a real-world application: optimizing email campaigns for fundraising. We aim to raise funds for a well-known humanitarian organization that supports various causes globally. The objective of our experiment is to send personalized emails to previous donors of the organization, encouraging them to make additional donations. While human experts have meticulously crafted various high-level email templates and strategies designed to maximize donation likelihood, it remains uncertain which specific type of email will resonate best with each individual donor. This uncertainty presents a challenge: aligning the email content with the unique preferences and motivations of each recipient to enhance the effectiveness of the campaign.

\paragraph{Users.} 
For this experiment, the ideal participants would be actual donors of our target charity. These users would be characterized by a rich set of attributes such as age, gender, location, occupation, hobbies, and their history of charitable donations. However, obtaining such comprehensive and detailed user data is not only challenging due to the sensitivity and privacy concerns associated with handling personal information but also financially and operationally demanding. Therefore, in this set of experiments, we propose an innovative alternative. Instead of relying on actual user datasets, we utilize a larger model's responses as an oracle to pre-train our bandit on the responses of a smaller model. In this experiment, we rely on OpenAI GPT-4o model \cite{achiam2023gpt} as a surrogate to the real potential donors, and simulate their preferences.

First, using GPT-4o, we create a diverse set of 1,000 virtual users. Each user is defined through a detailed textual profile, specifying attributes such as age, gender, location, occupation, hobbies, financial situation, motivations for donating to charity, and a history of past charitable contributions. Next, for each generated user $u_t$, we prompt GPT-4o to adopt their persona and simulate how they might interact with our contextual bandit. We embed the textual representation of each user using OpenAI's text embedding to define the context vector $\phi_t$. 

\paragraph{Arms.} We consider $K = 4$ arms. For each user, we prompt $\llm$ to generate different emails for each user (given the textual representation of the user) using one of the following styles: \textit{Formal}, \textit{Emotional/Narrative}, \textit{Informative/Educational}, and \textit{Personal/Relatable}. The detailed description of each instruction is in \Cref{sec:appendix-experimental-details}. All emails for each user are customized to match the characteristics of that user. For instance, if user $u_t$ is identified as a nature enthusiast, the content of emails may center around an environmental cause supported by the charity, crafted in different high-level styles of writing to cater to different communication preferences.

\paragraph{Rewards.} Rewards are generated according to \Cref{alg:CBLI} with each model $\llm$. We repeat the process five times and normalize the rewards. To quantify the regret, we use GPT-4o (which was our surrogate for real human users) as our oracle to generate the true preferences of each user. The oracle, representing an ideal decision-maker, selects the arm corresponding to the user's highest true reward.  

\paragraph{Baselines.} 
 We pre-train several contextual bandit models with the data generated according to \Cref{alg:CBLI}. Each experiment uses a different language model $\llm$. We also establish three additional baselines for comparison. We pre-train a \emph{Similarity} baseline, which considers the rewards of pulling each arm as the cosine similarity of the user's context vector and the content vector of each arm. \emph{Not pretrained} baseline illustrates the regret of a contextual bandit which is not pre-trained beforehand. Finally, \emph{GPT-4o (oracle)} baseline demonstrates a contextual bandit which is pre-trained with the rewards from the oracle.  

\begin{figure}[t]
\centering
\includegraphics[trim=0.5cm 0cm 1.5cm 1.5cm, width=\linewidth]{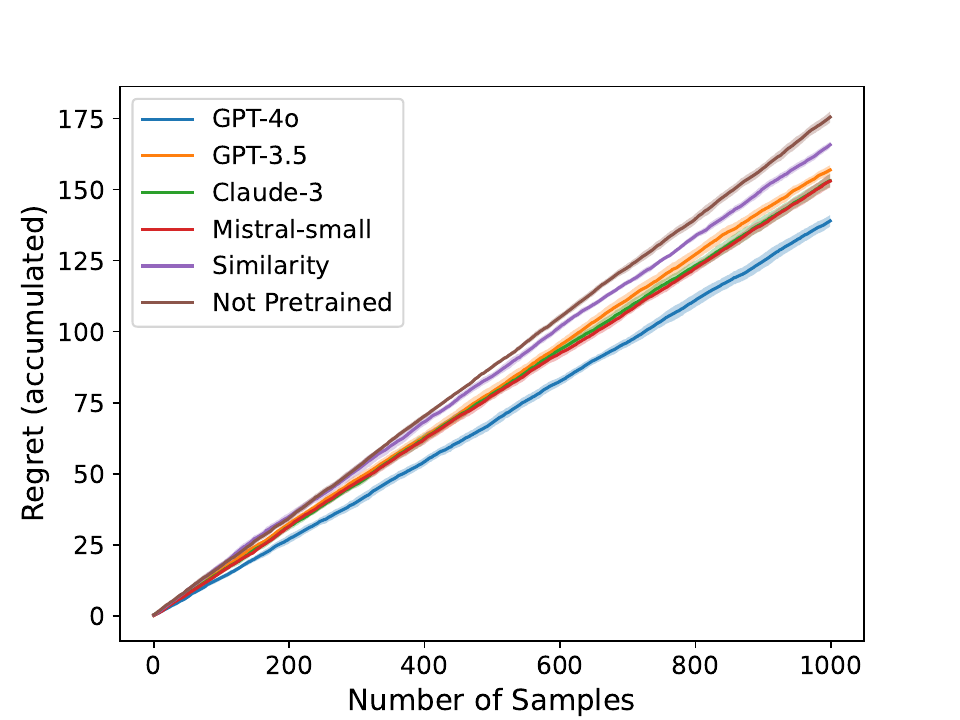}
\caption{Accumulated regret relative to a GPT-4o-based oracle across 1,000 samples. Each line represents a CB trained using data generated by CBLI with $\llm$ indicated in the legend. Error bars represent variance over shuffling true responses 10 times.}
    \label{figs:synthetic}
\end{figure}

\paragraph{Results.} 
We fine-tune each contextual bandit baseline for $T = 1,000$ steps with the true rewards of the users (generated by the oracle). 
\Cref{figs:synthetic} presents the accumulated regret for each baseline model over the course of fine-tuning. Regret is calculated relative to an oracle, which, at every step $t \in [T]$, selects the arm that yields the highest reward for the user. 
The results demonstrate a clear trend: models pre-trained with various LLMs (GPT-3.5, Claude-3 Haiku, and Mistral-small), show significantly lower accumulated regret compared to the baselines, including a model with no pre-training and a model pre-trained with cosine-similarity rewards (the Similarity baseline). Notably, the accumulated regrets of these models are remarkably close to the pre-trained model trained with the oracle's rewards. This proximity in performance confirms our claim that using LLM-generated users and rewards to train contextual bandits can substantially reduce regret and serve as a foundation for fine-tuning with actual user rewards.

\Cref{table:synthetic} illustrates the reduction in cumulative regret after $T = 1000$ steps for each model when pre-trained with varying amounts of CBLI-generated data. We evaluated the performance of the bandits after pre-training them with datasets containing 100, 500, and 1000 CBLI-generated users. As shown in the table, by increasing the number of CBLI-generated users, we find a greater reduction in cumulative regret, though we also observe substantially diminishing returns.

\begin{table}[t]
  \centering
  \resizebox{\columnwidth}{!}{\begin{tabular}{|c|c|c|c|}
   \hline
   \multirow{2}{*}{$\llm$} & \multicolumn{3}{c|}{Users} \\ \cline{2-4}
   & 100 & 500 & 1000 \\ \hline
   GPT-4o & $9.21\% \pm 0.25$ & $22.56\% \pm 0.21$ & $25.75\% \pm 0.20$ \\ \hline
   GPT-3.5 & $4.80\% \pm 0.22$ & $12.40\% \pm 0.32$ & $14.28\% \pm 0.13$ \\ \hline
   Claude-3 & $5.38\% \pm 0.28$ & $13.28\% \pm 0.31$ & $14.70\% \pm 0.26$ \\ \hline
   Mistral-small & $5.93\% \pm 0.30$ & $13.15\% \pm 0.23$ & $16.96\% \pm 0.23$ \\ \hline
   Similarity & $3.66\% \pm 0.30$ & $5.43\% \pm 0.26$ & $6.42\% \pm 0.29$ \\ \hline
  \end{tabular}}
  \caption{The reduction in cumulative regret after $T = 1000$ steps when pretraining a bandit with 100, 500, or 1000 CBLI-generated users and their rewards (with input LLM $\llm$) compared to CB with no pretraining. Note the oracle is GPT-4o.}
  \label{table:synthetic}

 \end{table}

\subsection{Second Experiment: A Choice-Based Conjoint Analysis with Real-world Data}\label{sec:experiments-conjoint}

We further extend our experiments by utilizing the results of a conjoint survey experiment. These experimental designs, which are popular in political science \cite{Bansak2021}, health \cite{trapero2019attributes}, and market research \cite{cattin1982commercial, louviere1994conjoint}, are designed to detect how citizens, patients, and customers value different characteristics of a candidate for public office, a health intervention, a product, or another object of interest. These objects usually have multiple features which may be causally linked to preferences. A conjoint survey proceeds by showing a participant descriptions of two or more objects and asks them to rank them according to some criterion, e.g., likelihood to purchase. This setup is an ideal scenario to test CBLI as these survey experiments capture both user demographic information, which can be used to generate contexts, as well as information on human preferences, which can be the target of the contextual bandit.

We focus on the conjoint survey experiment described in \cite{kreps2020factors} and the dataset \cite{DVN/6BSJYP_2020}. This survey, conducted before the widespread availability of COVID-19 vaccines, attempted to determine which properties of a hypothetical vaccine would lead to wider acceptance of the vaccine.

\paragraph{Users and setting.} In this study, $N = 1970$ participants begin by providing demographic and personal information such as age, gender, location, income range, religion, and political views. We use this information to construct a context vector for each user. Next, participants are presented with a pair of hypothetical COVID-19 vaccines and asked to express which they are more likely to accept. Each vaccine is characterized by a specific set of eight attributes, including the vaccine’s efficacy rate, the duration of protection it offers, and potential side effects, resulting in 576 possible descriptions. This process is repeated a total of five times per participant.\footnote{A full description of the dataset can be found on the \href{https://dataverse.harvard.edu/file.xhtml?fileId=4067062&version=2.0}{Harvard Dataverse}.}

To train our model according to \Cref{alg:CBLI}, we generate 10,000 synthetic users, based on the set of features of the actual users. We use these users to pre-train our contextual bandit.

Given the large number of potential vaccine combinations, we adapt our approach to the experience collection for CBLI, by not considering every possible vaccine as an individual arm. Instead, we assume that each vaccine, described by its set of features, is sampled from a population $\mathcal{V}$, where $|\mathcal{V}| = 576$. At each step $t \in T$, user $u_t$ is randomly selected from the generated users. Subsequently, we sample a pair of vaccines $v_1, v_2 \in \mathcal{V}$. Rather than treating each vaccine as an arm, we focus on the comparison between pairs of vaccines, which aligns more closely with how choices are presented in the conjoint analysis.

In terms of the contextual bandit model, a feature map, $\psi\!\!:\! \mathcal{C}\!\times\!\mathcal{V}\!\times\!\mathcal{V}\!\rightarrow\! \mathbb{R}^d$, jointly encodes the user context $\phi_i \in \mathcal{C}$ and the two vaccines $v_1 \in \mathcal{V}$ and $v_2 \in \mathcal{V}$, and then learns a linear model on $\psi$. 
In our experiments, we take $\psi(\phi_i, v_1, v_2) = \operatorname{flat}(\phi_i (f(v_1) - f(v_2))^T) $, where $\operatorname{flat}$ represents flattening a matrix of size $n \times m$ as a vector of size $nm$, and $f(v)$, $f\!\!:\!\mathcal{V}\!\rightarrow \mathbb{R}^7$, outputs the $7$ features of each vaccine in the dataset.

\paragraph{Arms} 
Here, we adopt the \emph{Sleeping Bandit} framework \cite{kleinberg2010regret} described in \Cref{sec:prelim} and \Cref{sec:method}. In our setup, two arms are available at each step $t$, corresponding to the pair of vaccines presented in the context when the conjoint experiment data was collected. Thus, at each decision point, the model must choose between only these two options.

\paragraph{Rewards.} This setup uses the sparse mode in \Cref{alg:CBLI}. For pre-training bandit data collection, for the pair of vaccines $i,j$, if the LLM-simulated user prefers $i$ over $j$, then $R[i,j] +\!\!=1$. $R_t$ for online learning from actual rewards at round $t$ works similarly but is based on real human preferences from the conjoint experiment log.

\paragraph{Results.}  

\begin{figure}[t]
\centering
\includegraphics[trim=0.5cm 0cm 1.5cm 1.25cm, width=\linewidth]{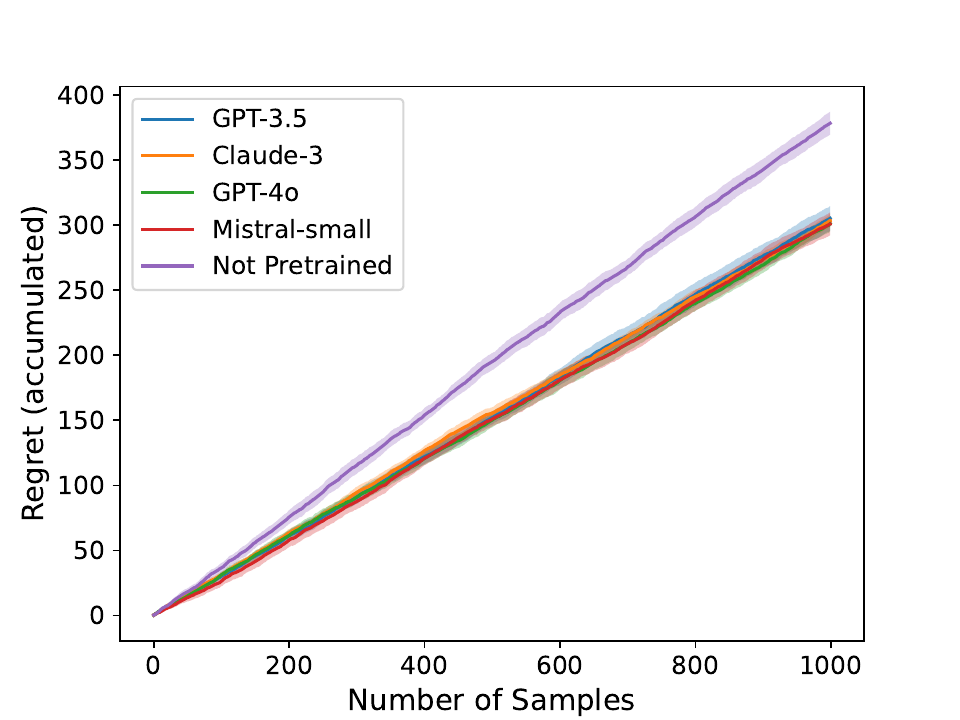}
\caption{Accumulated regret relative to true responses from $N=1970$ responses to a conjoint experiment. Each line represents a different instantiation of CBLI with a different LLM. Error bars represent variance over shuffling true responses 10 times.}
    \label{figs:conjoint}
\end{figure}

We demonstrate the impact of pre-training contextual bandits using data generated by LLMs in \Cref{figs:conjoint}. In our experiment, we configure the LLM $\llm$ with different models and pre-train the bandit using data from $10,000$ simulated users and their generated preferences. Once pre-trained, we deploy the bandit on the real users participating in the study and further fine-tune it using their actual responses. \Cref{figs:conjoint} presents the regret outcomes for various models when applied to true user data. Regret is computed based on the discrepancy between the model's recommendations and the actual preferences of the study participants. We compare these results against a baseline model, denoted as \emph{Not Pre-trained}, which is trained exclusively on the real user data without any pre-training. The results clearly show that models pre-trained with LLM-generated data exhibit significantly lower regret than the baseline that was not pre-trained. This demonstrates that using LLMs to simulate user preferences and pre-train the bandit models can effectively reduce the learning curve and improve performance when transitioning to actual user interactions.

Similar to \Cref{table:synthetic}, in \Cref{table:conjoint-2}, we explore the effect of differing amounts of pre-training data. We see that, for this example, more CBLI-generated data increases the performance of the bandits, though we observe sharply diminishing returns. The reduction in cumulative regret is measured after $T = 1000$ steps of fine-tuning. 

\begin{table}[t!]
  \centering
  \small
  \resizebox{\columnwidth}{!}{\begin{tabular}{|c|c|c|c|}
   \hline
   \multirow{2}{*}{$\llm$} & \multicolumn{3}{c|}{Users} \\ \cline{2-4}
   & 1000 & 5000 & 10000 \\ \hline
   GPT-4o & $17.10\% \pm 0.63$ & $16.07\% \pm 0.39$& $20.28\% \pm 0.41$  \\ \hline
   GPT-3.5 & $11.95\% \pm 0.67$ &  $14.32\% \pm 0.38$ & $19.19\% \pm 0.53$ \\ \hline
   Claude-3 & $13.79\% \pm 0.58$ & $17.18\% \pm 0.16$ & $19.65\% \pm 0.37$  \\ \hline
   Mistral-small & $14.51\% \pm 0.77$ & $17.89 \% \pm 0.40$ &  $20.39\% \pm 0.38$\\ \hline
  \end{tabular}}
  \caption{The reduction in cumulative regret of $T=1000$ steps when pretraining a bandit with 100, 1,000, or 10,000 CBLI-generated users and their rewards (with input LLM $\llm$) compared to CB with no pretraining.}
  \label{table:conjoint-2}
 \end{table}

\paragraph{Initializing Bandit with Partial Information}
To explore the robustness of our framework, we conduct additional experiments to evaluate the impact of varying amounts of user information on the effectiveness of pre-training the contextual bandit. We experiment with three distinct subsets of user data. \emph{No personal} baseline excludes any personal or demographic details about the users. It focuses solely on COVID-19 and vaccine-related information of each user.  \emph{Only personal} baseline includes only the personal and demographic information of each user, such as age, gender, location, and income range. It omits any data related to COVID-19 or vaccines about each user.  \emph{Partial personal} baseline only contains a reduced set of personal and demographic details compared to \emph{Only Personal} baseline. It includes a minimal amount of personal information, providing a middle ground between having information about user profiles and having none at all. \emph{Full} baseline includes all the information about the users. For each baseline, we compute the total accumulated regret after $T = 1000, 2000$ and $9855$ steps, using the responses of true users in the survey. $T= 9855$ steps is fine-tuning the model with all the data in the survey.
We begin by pre-training the bandit models using data from 10,000 generated users. All baselines are pre-trained with GPT-3.5 generated preferences of users. After pre-training, we fine-tune each bandit model with data from actual users, using their survey responses. To evaluate the effectiveness of each approach, we calculate the accumulated regret for each baseline model. The reduction in cumulative  regret is measured with respect to the regret of a bandit model that was not pre-trained and was only trained on real user data. The results, summarized in \Cref{table:conjoint} demonstrate that incorporating even minimal user context during pre-training can significantly enhance the performance of contextual bandit models. 
Furthermore, as the number of fine-tuning steps $T$ increases, the reduction in cumulative regret for each pre-trained baseline gradually diminishes. This convergence occurs because, over time, the contextual bandit model accumulates sufficient data from real users during the fine-tuning phase. As a result, the performance gap between the pre-trained models and the model that was not pre-trained becomes less pronounced.

\begin{table}
  \centering

  \resizebox{\columnwidth}{!}{\begin{tabular}{|c|c|c|c|}
   \hline
   \multirow{2}{*}{Data} & \multicolumn{3}{c|}{T} \\ \cline{2-4}
   & 1000 & 2000 & 9855 (all) \\ \hline
   Full & $19.19\% \pm 0.53$ &  $14.90 \% \pm 0.44$ & $8.07 \% \pm 0.22$\\ \hline
   No Personal & $14.79\% \pm 0.41$ & $11.60\% \pm 0.47$ &  $5.23\% \pm 0.15$  \\ \hline
   Partial Personal & $18.55\% \pm 0.51$ & $14.81\% \pm 0.46$ & $7.70\% \pm 0.19$ \\ \hline
   Only Personal &  $15.25\% \pm 0.45$ & $11.60\% \pm 0.47$ & $7.69\% \pm 0.20$ \\ \hline
  \end{tabular}}
\caption{The reduction in accumulated regret when pre-training contextual bandit models with varying levels of user information. Each row represents a different baseline, with different amount of user-specific context. }
\label{table:conjoint}
\end{table}

\section{Conclusion}
We introduced a novel framework, Contextual Bandits with LLM Initialization (CBLI), for jump-starting contextual bandits with Large Language Models (LLMs), aimed at enhancing early decision-making in personalized systems through pre-training on synthetic data. Our experiments demonstrated the potential of LLMs to generate a dataset that, while not perfectly mirroring true human preferences, provides a significantly better starting point than the traditional cold-start scenario in bandit models.
Through rigorous testing in two different bandit settings, we have shown that CBLI can reduce early regret by significant margins, indicating its practical effectiveness and robustness. 

\section{Limitations}
\label{sec:limitations}

While this work demonstrates that the output of LLMs can be used to jump-start a contextual bandit in some situations, we emphasize that in this work, we have focused on total, accumulated regret, which represents an improvement {\em on average}. Researchers have demonstrated that the output from LLMs can be biased in many distinct ways, perhaps most strikingly with respect to political ideology \cite{perez2023discovering, hartmann2023political}. Moreover, the distributions generated by LLMs heavily depend on the prompting strategy \cite{santurkar2023whose}. Indeed, some researchers have begun to question if the measures of fairness and techniques to make models fairer that have been developed in the ML community can even be applied to LLMs \cite{anthis2024impossibility}. At a high level, further work should be done examining the differential impact on regret for certain subpopulations of interest across a wide variety of tasks. At an individual project level, practitioners considering utilizing CBLI in a real-world system should be aware of these potential biases and consider their potential impacts on users.

Second, as noted above, we expect the distributions of rewards generated by LLMs to differ from the ground truth. While not observed in our studies, it is worth noting that, in theory, distributional misalignment can cause {\em worse} regret than cold starting the CB \cite{zhang2019warm}. Robustness techniques in prior work \cite{zhang2019warm} can be incorporated into CBLI to maximize its usefulness in the future.

Third, while well-designed, targeted communications have been utilized to encourage pro-social outcomes in a wide variety of contexts \cite{dellavigna2022rcts}, the same techniques can be utilized by malicious actors to anti-social ends. For instance, these tools could be used, for example, to create disinformation campaigns, to stoke negative sentiment toward outgroups, or to drive antipathy toward life-saving public health measures. These problems are true of any contextual bandit and especially true when paired with an LLM, though CBLI may jump-start any such effort.

Finally, our reward function estimation focuses on determining the rank order of rewards and not the {\em magnitude} of rewards. This is sufficient in traditional CB problems where the goal is to {\em maximize} total rewards, but may not be sufficient in other contexts, such as when CBs are used in adaptive treatment assignment with other goals or constraints \cite{bastani2021mostly, kasy2021adaptive}.

\section*{Acknowledgments}

The authors would like to thank Jake Bowers, David Glick, and Nathaniel Rabb for suggesting several potential conjoint survey experiments to explore, and Lorne Schell for helpful comments on an early version of this work.

\bibliography{jump-bandit}

\appendix
\newpage

\section{Pairwise comparisons yield approximate rank orders}
\label{sec:appendix_rank_order}

In \Cref{alg:CBLI}, for each user $u$ and each pair of arms $(k_1, k_2)$, we prompt an LLM $\llm$ to rank the arms. If it ranks $k_1$ higher than $k_2$, we say $R_u[k_1] = R_u[k_1] + 1$. Otherwise, we say $R_u[k_2] = R_u[k_2] + 1$. In this section, we show under certain assumptions on the true rewards, that the values $R_u$ will represent a rank order of the user's preferences.

We assume that the reward distribution for user $u$ and arm $i$ is $Y_i = \mathrm{Bern}(p_i)$ a Bernoulli random variable with probability $p_i$ of realizing $1$ and probability $1 - p_i$ of realizing $0$. If $p_i > p_j$, then, on average, the user $u$ will derive more reward from being assigned arm $i$ than arm $j$.

Consider the random variables \[ 
    Z_{ij} = \begin{cases}
        1 & Y_i > Y_j \\
        0 & Y_i < Y_j \\
        B & Y_i = Y_j
    \end{cases}
\]
where $B = \mathrm{Bern}(0.5)$ is a Bernoulli random variable with probability $0.5$ of being $1$ and probability $0.5$ of being $0$. Further, let $Z_i = \sum_j Z_{ij}$. Then, assuming the LLM $\llm$ faithfully represents the user's preferences, by the law of large numbers, we expect the unnormalized $R_u[i] \approx m \expectedvalue{\left[Z_i\right]}$ where $m$ is the number of iterations.

Now 
\begin{align*}
    \expectedvalue{\left[Z_{ij}\right]} 
        &= \Pr(Y_i > Y_j) + \frac{1}{2} \cdot \Pr(Y_i = Y_j) \\
        &= p_i (1 - p_j) + \frac{1}{2}\left[p_ip_j + (1 - p_i)(1 - p_j)\right] \\
        &= \frac{1}{2} (p_i - p_j + 1).
\end{align*}

Then $\expectedvalue{\left[Z_i\right]} = \frac{K}{2}(p_i + 1) - \frac{1}{2}\sum_j p_j$. Rewriting, we see that $\expectedvalue{\left[Z_i\right]} = \frac{K}{2}p_i + C$ where $C$ is a constant for all $i$. Thus, the $\expectedvalue{\left[Z_i\right]}$ have the same order as the $p_i$.

\section{Comparing Arm Scoring Methods with LLMs}
\label{app:scoring}
\begin{figure}[t]
\centering
\includegraphics[trim=5cm 2cm 4cm 2cm, width=0.88\columnwidth]{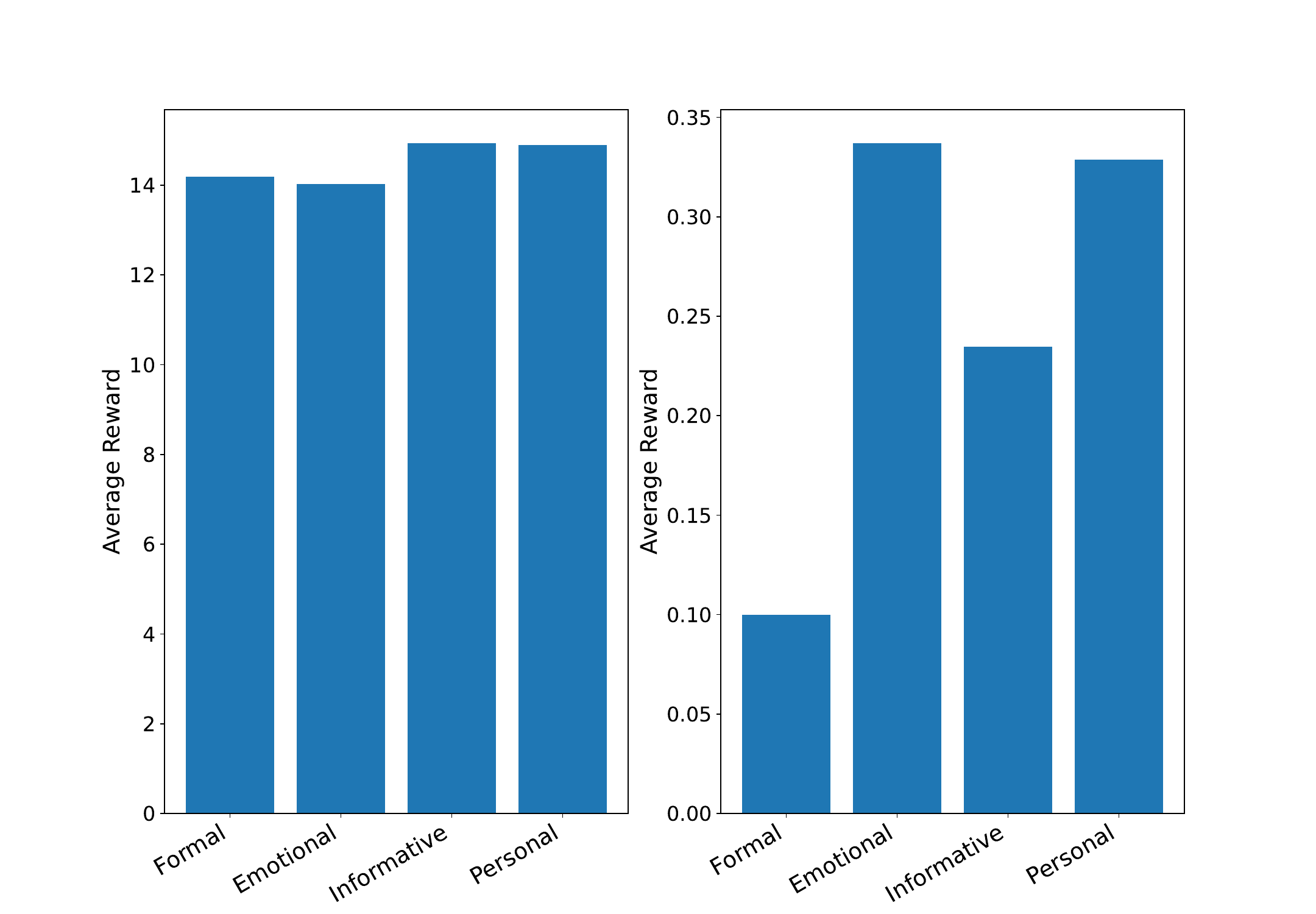}

\vspace{1\baselineskip}
\caption{Comparison of arm scoring methods using LLMs. Each bar represents the average reward of an arm across all users. The left figure illustrates results from scoring each arm individually, while the right figure shows results from scoring arms pair-wise as per \Cref{alg:CBLI}.}
    \label{figs:comparison}
\end{figure}

In our CBLI approach, outlined in \Cref{alg:CBLI}, we prompt an LLM $\llm$ to rank \emph{pairs} of arms as opposed to scoring arms individually, and determine an ordering based on them. We find that prompting $\llm$ with pairs of arms leads to more consistent results compared to scoring each arm independently. \Cref{figs:comparison} displays the average reward for each arm across all generated users using two different prompting methods. The left figure depicts the approach where $\llm$ is prompted for each user $i$ to provide a score between 0 and 100 for each arm $a$, proportional to the likelihood of that user donating. The rewards generated by this method show minimal variation and are difficult to distinguish on the scale of 0 to 100.
In contrast, the right figure illustrates the normalized rewards of each arm using the pair-wise prompting method. This approach captures the diverse preferences of users more effectively, revealing distinct patterns in user preferences across different arms. 

\section{Experimental Details}

In our experiments, we train LinUCB \cite{chu2011contextual} with $\alpha = 10$ as the hyperparameter. In all figures and tables, results are reported based on the average of 10 runs. 

We collect the data for our experiments using the LLM API provided by OpenAI, Anthropic, and Mistral \cite{achiam2023gpt, claude3, jiang2023mistral}. 

We train our bandit models on a system with the following specification: 2.3 GHz Quad-Core Intel Core i7 and 32 GB of RAM.
The total running time to train the models is less than 2 hours.

\label{sec:appendix-experimental-details}

\subsection{First Experiment: Personalized Email Campaign for Charity Donations}

In this set of experiments, we pre-train each bandit model with the dataset composed of size $n = 20 \times \text{Users}$ where the number of users is specified in \Cref{table:synthetic}. For each specified number of users, we randomly sample 
$n$ instances from the available user data to create the training dataset.

\paragraph{Arms.}
We consider $K = 4$ arms, correspond to the following styles: \textit{Formal}, \textit{Emotional/Narrative}, \textit{Informative/Educational}, and \textit{Personal/Relatable}.
For each user we generate a customized message based on each of these styles. Description of each style is as follows. 

\begin{itemize}
    \item   Formal Approach: Begin with a formal greeting, introduce the organization, highlight the current need, explain how their contribution can make a difference, and end with a polite request asking for their generous support. The tone is official and respectful, focusing on the importance of the cause.
    \item Emotional/Narrative Style: This style leverages storytelling to evoke empathy and compassion from the reader. You can share a real-life story related to the cause, emphasize the struggle, and showcase how their donation can change lives.
    
    \item Informative/Educational Style: This style relies on facts, statistics, and evidence to persuade the user to donate. It educates the reader about the cause, its impact, and how the charity is fighting for it. The reader's decision will be driven by the evidence of how efficient the charity work is.
    \item Personal/Relatable Style: Here, you use a more casual, friendly tone. You could even share personal experiences with the charity or testimonies from donors. The essence is to make the reader feel closely connected and to understand that anyone can make a difference.

\end{itemize}

\paragraph{Prompts. }
Given the information about each user $i$ and the style of the arm $k$, we generate a customized message for the user to be the content of arm $k$ with the following prompt:

\textit{Write a message for the following user to encourage them to donate to [name of charity] charity organization: [description of user $i$] Use the following style to generate the message: [description of style $k$]}

Furthermore, to generate the pair-wise rewards for user $i$ and arms $k_1, k_2$ as described in \Cref{alg:CBLI}, we use the following prompt:

\textit{Pretend you are the following user: [description of user $i$] Now you are receiving these two messages: [Message $k_1$] [Message $k_2$] Which message is more aligned with your interests? Which one makes you donate to the charity? Let's think step by step. Print the number of preferred message at the end: [Answer]}

\subsection{Second Experiment:  A Choice-Based Conjoint Analysis with Real-world Data}

In this set of experiments, we pre-train each bandit model with the dataset composed of size $n = 2 \times \text{Users}$ where the number of users is specified in \Cref{table:conjoint-2}. For each specified number of users, we randomly sample 
$n$ instances from the available user data to create the training dataset.

\paragraph{Full Description of Users.} Each respondent answered the following multi-choice questions: gender, race, age, state, income range, religious beliefs, political views, their (dis)approval of Donald Trump's handling of his job as a president,\footnote{As a reminder, this survey was taken in 2020.} recent changes in employment status, ability to work from home, whether they suffered from COVID, if they knew a person who had been hospitalized or had died because of COVID, whether (with respect to COVID) they believed the worst was behind us or yet to come, whether they had received a flu vaccination in the past, whether they think vaccines are safe in general, whether they think children should be required to be vaccinated against childhood diseases, and whether they have health insurance.

\paragraph{Full Description of Vaccines.} Each vaccine is described with the following attributes: efficacy, duration of protection,  major side effects, minor side effects, FDA approval, country of origin, and whether it is endorsed by a president or a health organization.

\paragraph{Prompts.} For each user $u_t$ and a pair of vaccines $v_1, v_2$, we prompt LLM with the following. 

\textit{Consider you are in the middle of the COVID pandemic where vaccines are just being produced. Pretend to be the following user: [description of $u_t$] now you are given two vaccine choices for COVID. The description of each vaccine is as follows: [description of $v_1$], now the next one: [description of $v_2$]. Which one do you take? A or B? Let's think step by step. Print the final answer as [Final Answer] at the end as well.}

\end{document}